\documentclass[conference]{IEEEtran}
\IEEEoverridecommandlockouts
\usepackage{cite}
\usepackage{amsmath,amssymb,amsfonts}
\usepackage{algorithmic}
\usepackage{graphicx}
\usepackage{textcomp}
\usepackage{xcolor}

\usepackage{microtype}
\usepackage{booktabs}
\usepackage{hyperref}
\usepackage{array}
\usepackage{multirow}
\usepackage{algorithm}
\usepackage{algorithmic}
\usepackage{comment}
\usepackage{subcaption}
\usepackage{wrapfig}

\def\BibTeX{{\rm B\kern-.05em{\sc i\kern-.025em b}\kern-.08em
    T\kern-.1667em\lower.7ex\hbox{E}\kern-.125emX}}
\begin{document}

% \title{{\fontsize{19.45}{0}\selectfont TRICE: Improving Worst-Case Performance of NVCiM DNN Accelerators through Training with Right-Censored Gaussian Noise}
% % \thanks{Identify applicable funding agency here. If none, delete this.}
% }

\title{{\fontsize{23}{0}\selectfont On the Viability of using LLMs for SW/HW Co-Design: An Example in Designing CiM DNN Accelerators}
}

% \author{\ }

\author{
    \textbf{
        Zheyu Yan\textsuperscript{$\ddagger$} \ \ \ \ 
        Yifan Qin \ \ \ \ 
        Xiaobo Sharon Hu \ \ \ \ 
        Yiyu Shi\textsuperscript{$\dagger$}
    }\\
    \IEEEauthorblockA{University of Notre Dame, \{\textsuperscript{$\ddagger$}zyan2, \textsuperscript{$\dagger$}yshi4\}@nd.edu}
    \vspace{-1cm}
}

\maketitle

\begin{abstract}
% DNN is good. We want DNN on edge.
% DNN on edge is hindered by power and computing budgets.
% SW-HW co-design solves this well.
% SW-HW co-design solves this slowly because of random initialization or ``cold start''.
% We use LLMs to address this.
% We achieve 25x speedup.

Deep Neural Networks (DNNs) have demonstrated impressive performance across a wide range of tasks. However, deploying DNNs on edge devices poses significant challenges due to stringent power and computational budgets. An effective solution to this issue is software-hardware (SW-HW) co-design, which allows for the tailored creation of DNN models and hardware architectures that optimally utilize available resources. However, SW-HW co-design traditionally suffers from slow optimization speeds because their optimizers do not make use of heuristic knowledge, also known as the ``cold start'' problem. In this study, we present a novel approach that leverages Large Language Models (LLMs) to address this issue. By utilizing the abundant knowledge of pre-trained LLMs in the co-design optimization process, we effectively bypass the cold start problem, substantially accelerating the design process. The proposed method achieves a significant speedup of 25x. This advancement paves the way for the rapid and efficient deployment of DNNs on edge devices.

\end{abstract}

% \begin{IEEEkeywords}
% component, formatting, style, styling, insert
% \end{IEEEkeywords}
\section{Introductions}

Deep neural networks (DNNs) have made extraordinary advancements, exceeding human performance across numerous perception tasks. The recent emergence of deep learning-based generative models, including DALL-E~\cite{ramesh2021zero} and the GPT family~\cite{brown2020language}, have significantly transformed our workflows. Currently, there is a clear trend toward embedding on-device intelligence into edge platforms such as mobile phones, watches, and cars, revolutionizing various aspects of daily life~\cite{chen2016eyeriss}. However, these edge platforms, with their limited computational resources and stringent power constraints, pose significant challenges in DNN deployment. These conditions call for the development of more energy-efficient DNN hardware, going beyond general-purpose CPUs and GPUs.

Application-Specific Integrated Circuits (ASICs) and Field-Programmable Gate Arrays (FPGAs) are potential solutions for achieving energy-efficient DNN inference on edge devices, given their ability to be specifically tailored to accelerate DNN operations. Additionally, Compute-in-Memory (CiM) DNN accelerators~\cite{shafiee2016isaac, yan2022swim, yan2022computing, yan2020single} emerge as strong contenders to replace CPUs and GPUs for DNN inference acceleration. Unlike traditional von Neumann architecture platforms, which necessitate frequent data transfers between memory and computational components, CiM DNN accelerators lower energy consumption by facilitating in-situ computation right at the data storage location. Furthermore, the advent of emerging non-volatile memory (NVM) devices, such as ferroelectric field-effect transistors (FeFETs) and resistive random-access memories (RRAMs), allows NVCiM accelerators to reach superior memory density and improved energy efficiency, surpassing conventional MOSFET-based designs~\cite{chen2016eyeriss}.

However, focusing solely on the hardware aspect is insufficient to attain high-efficiency DNN inference. The software aspect, \emph{i.e.}, DNN models, must also be specifically designed to optimize the underlying hardware. To address this, a hardware-aware Neural Architecture Search (NAS) approach~\cite{wu2019fbnet} has been proposed. This approach automatically explores DNN topologies and pinpoints the optimal design that balances high DNN performance with low hardware cost. Taking a step further, Software-Hardware (SW-HW) co-design of DNN accelerators~\cite{jiang2020hardware} has emerged, where DNN topology and underlying hardware are simultaneously designed. This results in superior performance and reduced hardware cost, since the optimal solution across both design spaces is co-explored, rather than being designed separately as in hardware-aware NAS. The NAS-based SW-HW co-design provides state-of-the-art performance in balancing between DNN performance and hardware cost.

However, NAS-based SW-HW co-design is immensely time-consuming, often requiring hundreds of GPU hours. This is due to the need to explore hundreds, if not thousands, of design candidates.

A significant contributor to this time consumption is the ``cold start'' issue. Regardless of whether Reinforcement Learning (RL) or Genetic Algorithms-based methods are used, the design optimizer starts from random guesses among all possible designs, gradually converging to the optimal solution. Essentially, the design optimizer learns the attributes of an optimal design from scratch. Yet, in reality, there is an abundance of heuristic knowledge that could be utilized in training the optimizer. For instance, given the same underlying hardware, a DNN model with more channels in each layer generally achieves higher accuracy, albeit at a higher hardware cost. Regrettably, such information cannot be used to train RL or Genetic Algorithm-based methods, as there is no clear way to generate rewards utilizing such information.

In response to this challenge, we propose to employ Large Language Models (LLMs) as the design optimizer\footnote{This research was partially supported by ACCESS 
% – AI Chip Center for Emerging Smart Systems 
, sponsored by InnoHK funding, Hong Kong SAR.}. LLMs, trained on extensive human language corpora, can comprehend content such as research papers and codes. They can, therefore, be fine-tuned by incorporating the latest research outcomes in DNN topology design, DNN accelerator design, and HW-SW co-design. In doing so, LLMs become specialists in these fields and can utilize the heuristics in the process of suggesting candidate designs, thereby circumventing the ``cold start'' issue.

In this work, we implement such a framework that adopts LLMs to perform SW-HW co-design and validate its efficiency via experiments. The contributions of this work are multifold:
\begin{itemize}
    \item We introduce LCDA, the first-ever framework that utilizes \underline{L}arge Language Models for the hardware-software \underline{C}o-\underline{D}esign of \underline{A}ccelerators for deep neural networks.
    \item We validate the efficacy of LCDA by co-designing the DNN topology and Compute-in-Memory (CiM) DNN accelerators.
    \item Our experimental results indicate a substantial 25X speedup compared to the state-of-the-art (SOTA) NACIM method~\cite{jiang2020device}, all while maintaining similar performance levels.
\end{itemize}
\section{Related Works}\label{sect:related}
\subsection{Neural Architecture Search and SW-HW Co-Design}
Neural Architecture Search (NAS) has emerged as a game-changer in various machine learning applications such as image classification, image segmentation, and video action recognition. By autonomously discovering high-performing neural architectures, NAS obviates the need for expert manual design. A standard NAS system encompasses a controller and a trainer. The controller generates neural architecture parameters, or \emph{child networks}, which are subsequently trained and assessed for accuracy. Feedback from this process informs updates to the controller. The search procedure continues until a pre-set limit of predicted child networks is achieved, with the highest accuracy architecture ultimately selected. While past research has demonstrated that automatically discovered neural architectures can rival the accuracy of those designed by humans, they often come with complex structures that can hinder their practical application, such as necessitating excessive bandwidth for secure inference.

In the specific domain of software-hardware (SW-HW) co-design, researchers~\cite{jiang2020device, wu2019fbnet} have advocated a co-exploration methodology that amalgamates the exploration of neural architecture and hardware design spaces. This approach, contrasting with those focused solely on fixed hardware, accommodates various hardware platforms, resulting in viable hardware solutions that deliver higher accuracy without breaching time constraints. Through this integrative approach, co-exploration enables the identification of highly efficient solutions. For instance, the authors of~\cite{jiang2019accuracy} present the FNAS framework, which concurrently determines the FPGA accelerator design and computation schedule during the child networks' search. Additionally, in~\cite{han2019design}, the authors expand the methodology to include quantization considerations during FPGA design, thereby further boosting system performance.

\subsection{Compute-in-Memory DNN Accelerators}\label{sec:2.1}

\begin{figure}[ht]
    % \vspace{-0.4cm}
    \centering
    \begin{minipage}[b]{0.53\linewidth}
        \includegraphics[trim=10 215 510 10, clip, width=1.\linewidth]{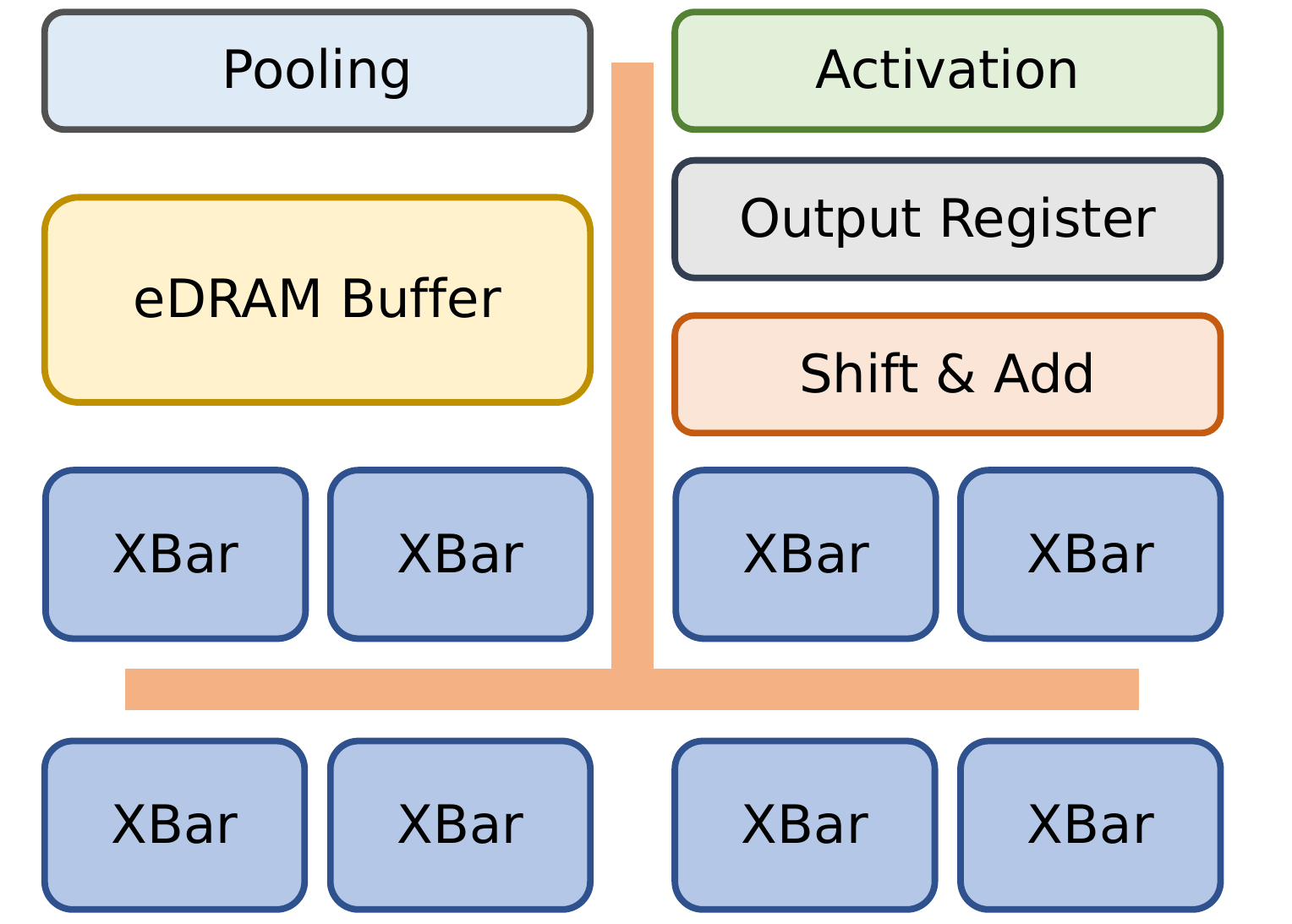}  
        \vspace{-0.5cm}
        \subcaption{CiM DNN accelerator.}
    \end{minipage}
    \begin{minipage}[b]{0.4\linewidth}
        \includegraphics[trim=0 150 550 0, clip, width=1.\linewidth]{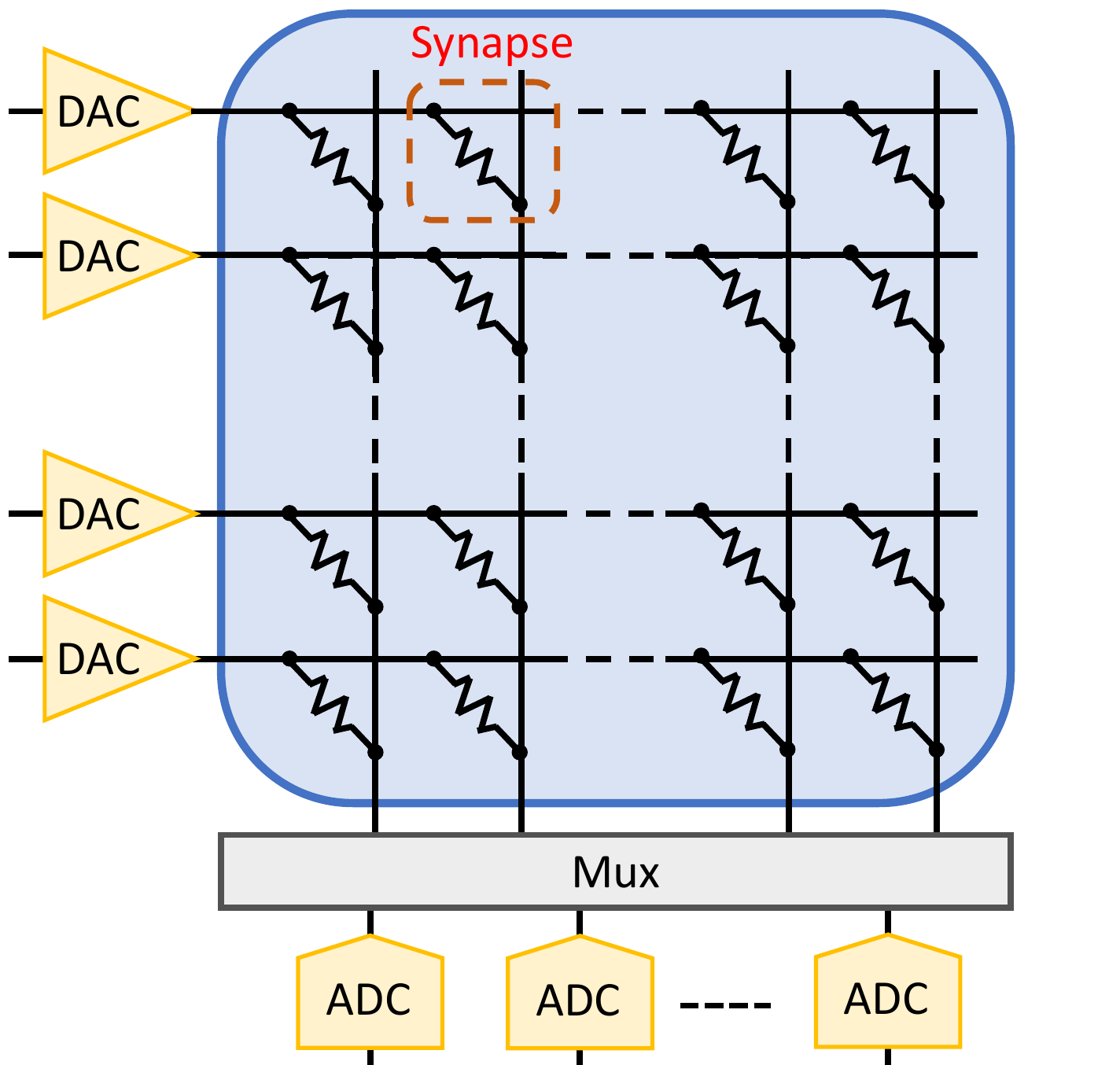}
        \vspace{-0.5cm}
        \subcaption{Crossbar array.}
    \end{minipage}
    \vspace{-0.1cm}
    \caption{Illustration of the NVCiM DNN accelerator architecture for (a) architecture overview and (b) crossbar (XBar) array. In a crossbar array, the input is fed horizontally and multiplied by weights stored in the NVM devices at each cross point. The multiplication results are summed up vertically and the sum serves as an output. The outputs are converted to the digital domain and further processed using digital units such as non-linear activation and pooling.}
    \vspace{-0.4cm}
\end{figure}

The core computational engine of NVM-based CiM DNN accelerators is the crossbar array structure. Capable of performing matrix-vector multiplication in a single clock cycle, crossbar arrays store matrix values (for example, weights in DNNs) at the intersecting points of vertical and horizontal lines using NVM devices such as RRAMs and FeFETs. Vector values, like inputs for DNNs, are introduced via horizontal data lines, or word lines, in voltage form. The resultant output is then conveyed through vertical lines (bit lines) in the form of current. While the crossbar array executes computations in the analog domain following Kirchhoff's laws, ancillary digital circuits are required for other fundamental DNN operations such as shift \& add, pooling, and non-linear activation. Intermediate data storage requires additional buffers, and digital-to-analog as well as analog-to-digital conversions are necessary between different domain components.

However, NVM-based crossbar arrays are prone to several variations and noise sources, including both spatial and temporal variations. Spatial variations stem from manufacturing defects and can manifest on both local and global scales. Moreover, NVM devices are subject to temporal variations resulting from random fluctuations in the device material. These variations in conductance can arise when the device is programmed at varying instances. Contrary to spatial variations, temporal variations are generally device-independent but may be influenced by the programmed value~\cite{feinberg2018making}. For this research, we consider the non-idealities to be uncorrelated amongst the NVM devices. Nonetheless, our framework can be adjusted to account for other variation sources with suitable modifications.

\section{Proposed Method}\label{sect:proposed}

In this section, we present LCDA, our novel framework utilizing \underline{L}arge \underline{L}anguage \underline{M}odels for the hardware-software \underline{C}o-\underline{D}esign of \underline{A}ccelerators for deep neural networks. LCDA seeks to identify the optimal pairing of DNN topology and hardware design within a user-provided design space, balancing high DNN performance with low hardware cost. Like existing co-design frameworks, LCDA comprises four core components: (1) design optimizer, (2) design generator, (3) DNN performance evaluator, and (4) hardware cost evaluator. In contrast to prior works, our approach innovatively incorporates large language models in the design optimizer, while the other components remain unchanged. To elucidate our proposed methodology, we use the co-design of DNN topology and Compute-in-Memory (CiM) DNN accelerators as an illustrative example, providing a detailed overview of the implementation of these four major components.

\subsection{Design Optimizer}

The design optimizer serves as a crucial component in automated hardware-software co-design. It processes all historical information about previously explored designs during the co-design process and generates a fresh DNN topology-hardware specification pair believed to deliver optimal performance and minimum hardware cost. 
Existing methods using reinforcement learning (RL)~\cite{jiang2020hardware}, or genetic algorithms~\cite{lu2019nsga}, have demonstrated effectiveness in generating such optimal design pairs. However, these methods necessitate the exploration of hundreds of diverse designs and are thus notably time-consuming, often requiring hundreds of GPU hours. 

This time consumption issue largely arises from the ``cold start'' problem. Whether utilizing RL or genetic algorithm-based methods, the design optimizer commences from a random selection among all possible designs and incrementally converges toward the optimal solution. Consequently, the design optimizer learns the characteristics of an optimal design from scratch. This process overlooks a wealth of heuristic knowledge that could be harnessed to train the optimizer.

To tackle this problem, we propose leveraging large language models (LLMs) as the design optimizer. LLMs, trained on extensive human language corpora, possess the ability to understand diverse content such as research papers and code. This allows them to be fine-tuned using recent research findings in DNN topology design, DNN accelerator design, and hardware-software co-design. Consequently, LLMs become proficient in these domains and can draw upon these heuristics during candidate design generation, thereby circumventing the ``cold start'' issue. In this work, we employ GPT-4~\cite{brown2020language}, pre-trained on the aforementioned content, as the design optimizer.

The application of LLMs necessitates meticulously crafted prompts (inputs) to elicit desired behaviors. We follow the approach demonstrated by GENIUS~\cite{zheng2023can}. An exemplary prompt template is presented in Algorithm~\ref{alg:prompts}.

\begin{algorithm}
    \caption{GPT-Prompts~($l_{des}$, $l_{perf}$, \textit{Model}, \textit{Choices})}
    \begin{algorithmic}\label{alg:prompts}
        \STATE // Input: a list of explored design $l_{des}$, the corresponding normalized performance of each design $l_{perf}$, design backbone \textit{Model}, and design space \textit{Choices}.
        \STATE // Output: a prompt to the GPT model;
        \STATE $prompt_s = $``\textit{You are an expert in the field of neural architecture search.}'';
        \STATE $prompt_u = $ ```\textit{Your task is to assist me in selecting the best rollout numbers for a given model architecture. The model will be trained and tested on CIFAR10, and your objective will be to maximize the model's performance on CIFAR10.
        The model architecture will be defined as the following.}
        \textit{\{Model\}}
        
        \textit{For the `rollout' variable to design the model, the available number for each index would be:} \textit{\{Choices\}}
        
        \textit{Your objective is to define the optimal number of rollouts for each layer based on the given options above to maximize the model's performance on CIFAR10. }
        
        \textit{The model's performance is a combination of hardware performance and model accuracy. If the hardware is invalid (e.g., too large in area), the performance I give you will be -1. After you give me a rollout list, I will give you the model's performance I calculated.}
        
        \textit{Your response should be the rollout list consisting of 6 number pairs(e.g. [[32,3],[32,3],[64,3],[64,3],[128,3],[128,3]]).}

        \textit{Here are some experimental results that you can use as a reference:} \{$l_{des}$, $l_{perf}$\}
        
        \textit{Please suggest a rollout list that can improve the model's performance on CIFAR10 beyond the experimental results provided above.}
        \textit{Please do not include anything else other than the rollout list in your response.}'''
        \STATE return $prompt_s$ + $prompt_u$;
    \end{algorithmic}
\end{algorithm}

\subsection{Design Generator}
The design generator derives a DNN topology and a hardware instance from each design candidate proposed by the design optimizer. For the parsing of GPT-4 outputs and subsequent generation of DNN topologies, we adopt the method presented by~\cite{zheng2023can}. We use the approach detailed in~\cite{jiang2020device} to generate hardware design specifications for Compute-in-Memory (CiM) DNN accelerators.

\subsection{DNN Performance Evaluator}

To evaluate the performance of the DNN topology produced by the design generator, the topology first needs to be trained using the target dataset, and then tested in the target environment. As DNN models deployed on CiM DNN accelerators are susceptible to the influence of device variations, we employ the noise injection training method~\cite{jiang2020device} for each DNN topology. The performance of the DNN under the effects of device variations is evaluated using the Monte Carlo simulation-based method~\cite{yan2021uncertainty}.

\subsection{Hardware Cost Evaluator}

Following the method in NACIM~\cite{jiang2020device}, we leverage the open-source simulation tool, DNN+NeuroSIM~\cite{peng2019dnn+}, to evaluate the hardware cost of each design candidate proposed by the design generator. DNN+NeuroSim is a comprehensive framework that simulates deep neural network (DNN) inference and on-chip training performance on hardware accelerators based on near-memory or in-memory computing architectures. It is compatible with various device technologies, including SRAM and emerging non-volatile memory (NVM) like RRAM, PCM, STT-MRAM, and FeFET. As a circuit-level macro model, it serves to benchmark neuro-inspired architectures by evaluating circuit-level performance metrics such as chip area, latency, dynamic energy, and leakage power. With PyTorch and TensorFlow wrappers, the DNN+NeuroSim framework supports a hierarchical organization from device-level properties to circuit-level modules, chip-level structures, and algorithm-level evaluation. 

\subsection{Framework Overview}

The LCDA framework executes up to a user-specified maximum of $EP$ episodes, exploring $EP$ design candidates in the process. In each iteration, a prompt is generated by Alg.~\ref{alg:prompts} that encompasses the design space description and results of past explorations, and this prompt is fed to the LLM model. The responses from the LLM model are parsed by the design generator to produce a DNN model and a hardware design instance. These outputs are subsequently assessed by the \textit{DNN Performance Evaluator} and \textit{Hardware Cost Evaluator} for performance and cost, respectively. The DNN model performance and hardware cost are consolidated into a single performance score through a reward function. This candidate design and its corresponding performance score are then documented for historical reference. Subsequently, LCDA advances to the next iteration.

\begin{algorithm}
    \caption{LCDA~(\textit{Model}, \textit{Choices}, $EP$, $f$)}
    \begin{algorithmic}\label{alg:lcda}
        \STATE // Input: DNN and hardware design backbone \textit{Model}, design space \textit{Choices}, number of maximum design choices to be explored $EP$, and reward function $f$;
        \STATE // Output: best design candidate $des_{opt}$
        \STATE Initialize design list $l_{des}$;
        \STATE Initialize normalized performance list $l_{perf}$;
        \FOR{i in 0...$EP$}
            \STATE $prompt = $ GPT-Prompts~($l_{des}$, $l_{perf}$, \textit{Model}, \textit{Choices});
            \STATE Send prompt to LLM;
            \STATE Parse LLM respond and get $des_i$;
            \STATE Generate DNN topology and hardware instance using \textit{Design Generator};
            \STATE Set \textit{DNN Performance Evaluator} result as $acc_i$;
            \STATE Set \textit{Hardware Cost Evaluator} result as $hw_i$;
            \STATE $perf_i = f(acc_i, hw_i)$;
            \STATE Add $des_i$ and $perf_i$ to $l_{des}$ and $l_{perf}$, respectively;
        \ENDFOR
        
    \end{algorithmic}
\end{algorithm}
\section{Experimental Evaluation}\label{sect:exp}
In this study, we employ LCDA to design the pairs of DNN models and CiM DNN accelerators, following the search space outlined in NACIM~\cite{jiang2020device}. We specifically concentrate on the design of DNN models for image classification tasks using the CIFAR-10~\cite{krizhevsky2009learning} dataset. The target DNN model consists of six convolution layers and two fully connected layers. LCDA is tasked with designing the kernel size and number of output channels, while the hidden size between the fully connected layers is set at 1024. 

The target CiM DNN accelerator design follows the ISAAC model~\cite{shafiee2016isaac}. LCDA determines the hyperparameters for hardware specification within the same search space as NACIM~\cite{jiang2020device}. Once the DNN models and hardware design hyperparameters are established, LCDA uses NeuroSIM to produce a CiM DNN accelerator tailored to this specific design.

In order to demonstrate LCDA's effectiveness, we perform two sets of experiments focusing on the multi-objective SW-HW co-design of CiM DNN accelerators. The first experiment seeks to strike a balance between DNN accuracy and energy consumption during inference, while the second targets a trade-off between DNN accuracy and inference latency. In both experiments, we compare the performance of LCDA with the state-of-the-art SW-HW co-design method, NACIM, which employs reinforcement learning as its optimization strategy.

\subsection{Design Trade-offs for DNN Accuracy and Energy}
We use a reward function of
\begin{equation}
    reward_{ae} = Accuracy - \sqrt{\frac{Energy}{8\times 10^7}}
\end{equation}
where $8\times 10^7$ is a normalization factor that normalizes the energy of each design to the original ISAAC design. Note that the energy is the lower the better.

\begin{figure}[ht]
    \vspace{-0.2cm}
    \centering
    \includegraphics[trim=00 0 0 0, clip, width=0.7\linewidth]{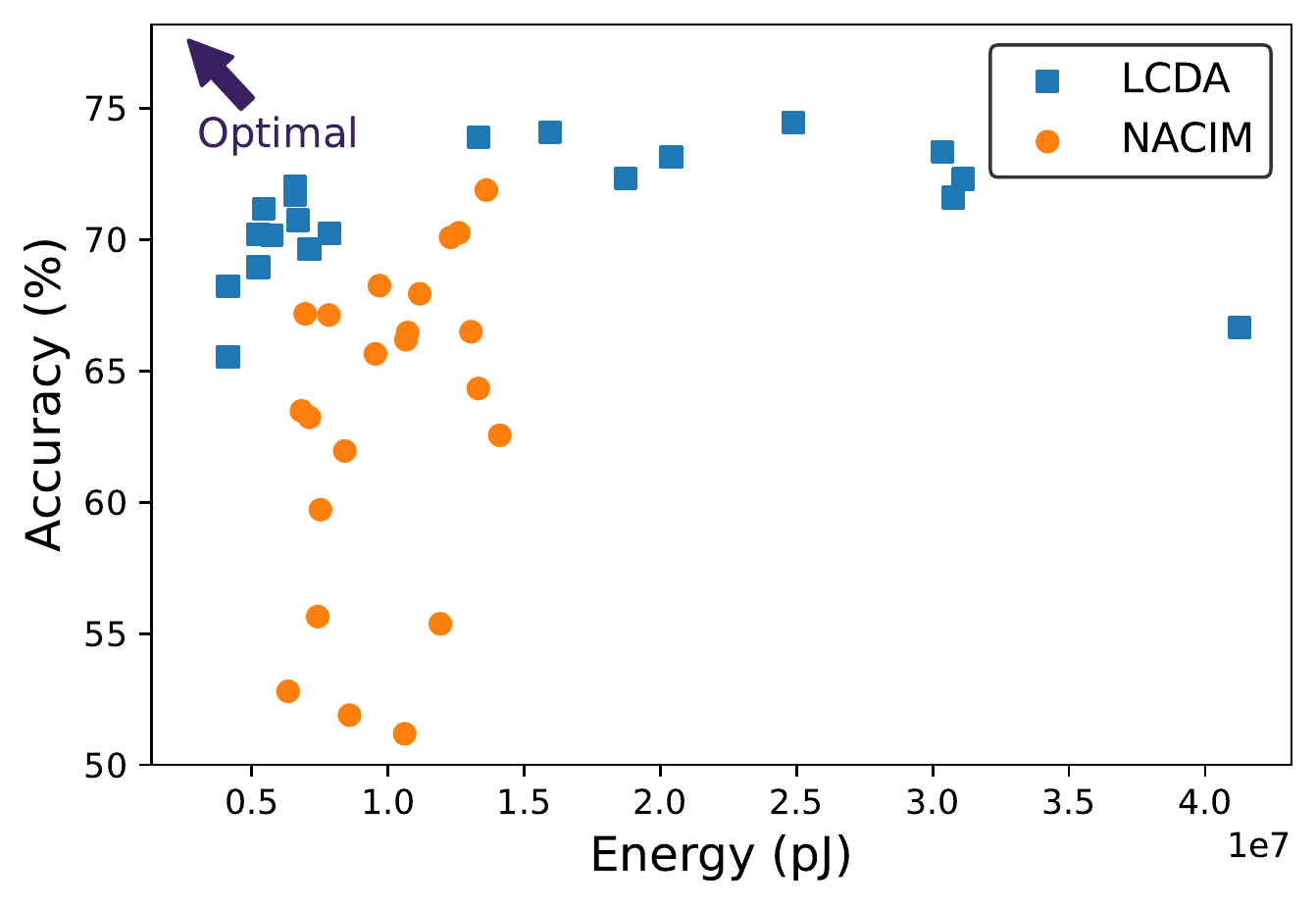}
    \vspace{-0.2cm}
    \caption{Accuracy-energy trade-offs of different design candidates provided by LCDA (blue square) and NACIM (orange dot). The Y-axis represents accuracy and the X-axis represents energy consumption in pJ.}
    \label{fig:sc_eng}
    \vspace{-0.2cm}
\end{figure}

In Figure~\ref{fig:sc_eng}, design candidates provided by both LCDA and NACIM are demonstrated. Each point in this figure denotes one design candidate. Its position on the Y-axis signifies the design's accuracy, while the X-axis indicates its energy consumption (in pJ). Designs located nearer to the upper-left corner are preferable. The blue dots correspond to design candidates from LCDA, while the orange squares are those from NACIM. When focusing on the upper-left region, LCDA and NACIM exhibit similar optimal results, and the Pareto Frontiers of both designs are alike. However, NACIM prioritizes candidates with lower energy consumption, leading to designs with somewhat diminished accuracy. Conversely, LCDA presents a spectrum of candidate designs with various energy consumptions, all yielding a reasonably high level of accuracy.

Digging deeper into each design candidate's parameters reveals varying strategies between LCDA and NACIM, even though both explore a range of hardware design parameters. NACIM typically maintains the channel size constant while exploring diverse kernel sizes, often leading to less desirable kernel sizes such as (1,7). It also ventures into designs with fewer output channels than input channels. In contrast, LCDA centers its exploration around different output channel numbers while always maintaining logical design choices. For instance, LCDA ensures that each layer's output channel number is greater than or equal to its input channel number, and never increases the number of output channels by more than 4 times.

Building on the knowledge of DNN properties, DNN accelerator characteristics, and SW-HW co-design principles, LLMs in LCDA can efficiently sidestep offering unreasonable design candidates. This significantly reduces the time and effort required for evaluating such candidates, leading to a markedly more efficient process of finding optimal designs. This efficiency is highlighted in a comparison of required episodes: while NACIM necessitates a minimum of 500 episodes (i.e., exploring 500 design candidates) to pinpoint the optimal solution, LCDA can unearth comparable solutions within just 20 episodes. This staggering difference translates into a speedup of 25 times, a crucial achievement that underscores the effectiveness and efficiency of the LCDA framework.

\begin{figure}[ht]
    \vspace{-0.2cm}
    \centering
    \includegraphics[trim=00 0 0 0, clip, width=\linewidth]{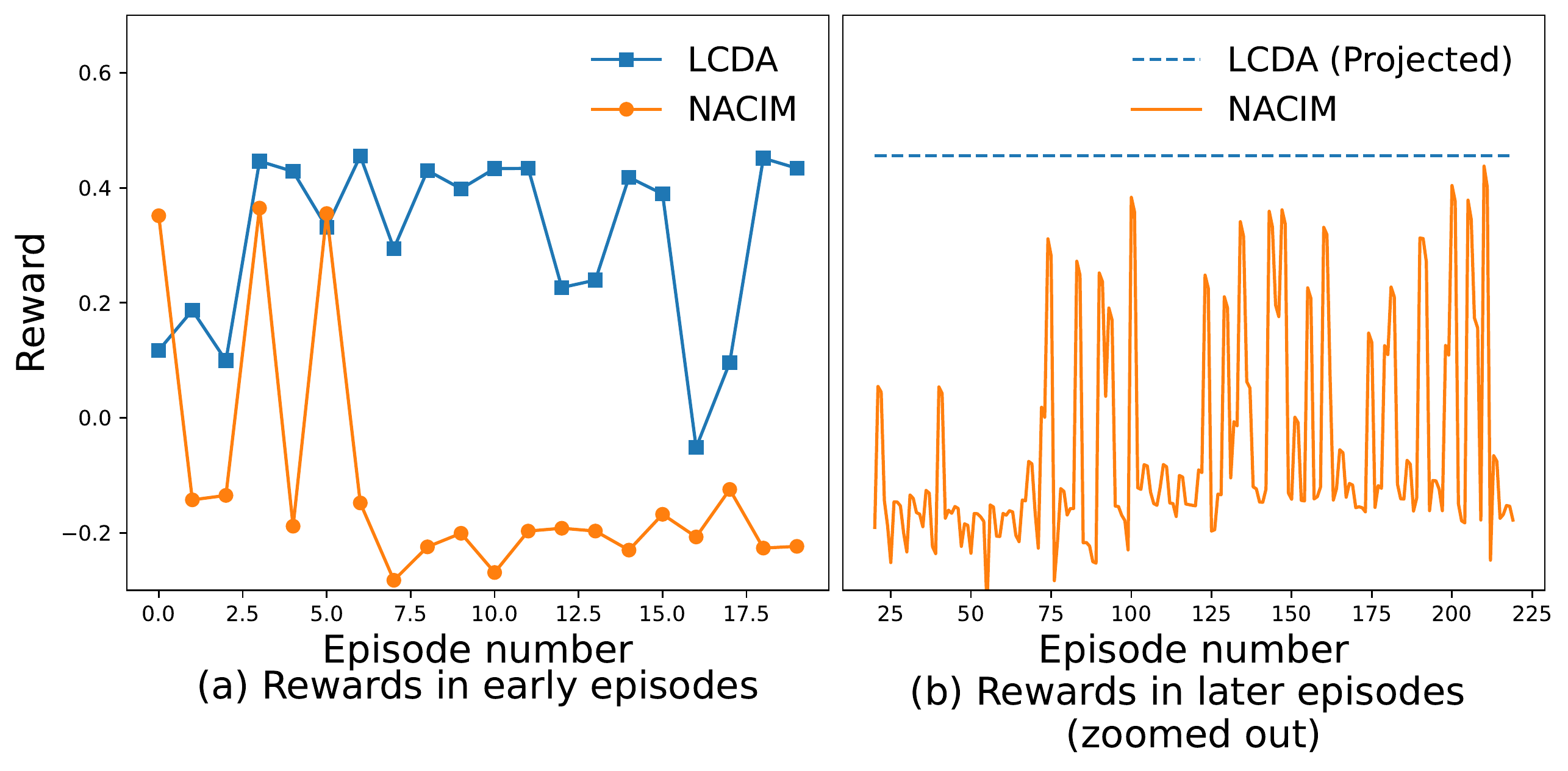}
    \vspace{-0.2cm}
    \caption{Rewards of different design candidates provided by LCDA (blue line) and NACIM (orange line). Figure (a) shows the results for the first 20 episodes and (b) shows the results of the $21^{st}$ to $500^{th}$ episode. Note that we only perform 20 episodes of search in LCDA, so we use the maximum reward of the first 20 episodes of LCDA to project its results in (b).}
    \label{fig:rwd_eng}
    \vspace{-0.2cm}
\end{figure}

To further emphasize this remarkable advantage, Figure~\ref{fig:rwd_eng} presents the reward values of design candidates provided by both methods across different search episodes. Notably, in LCDA, we only conduct 20 search episodes, and so we project the maximum reward from LCDA's first 20 episodes into later episodes in Figure~\ref{fig:rwd_eng} (b). Both NACIM and LCDA start with designs that receive a high reward. LCDA consistently explores designs with high rewards, while NACIM follows a more random approach. In later episodes, NACIM gradually approaches LCDA's reward values, indicating that it's slowly learning the knowledge that LCDA has from the get-go. This serves to further highlight that with prior SW-HW co-design knowledge, LCDA effectively bypasses the 'cold start' problem typical of the RL-based NAS method that NACIM uses, leading to a remarkable speedup. This 25-fold increase in efficiency underlines the significant advantage offered by the LCDA framework and its incorporation of large language models.

\subsection{Design Trade-offs for DNN Accuracy and Latency}
We use a reward function of
\begin{equation}
    reward_{al} = Accuracy + \frac{1}{Latency} \times \frac{1}{1600}
\end{equation}
where $\frac{1}{Latency}$ indicated frame per second (FPS) and $1600$ FPS is a normalization factor that normalizes the FPS of each design to the original ISAAC design. Note that the latency is the lower the better.

\begin{figure}[ht]
    \vspace{-0.2cm}
    \centering
    \includegraphics[trim=0 0 0 0, clip, width=0.7\linewidth]{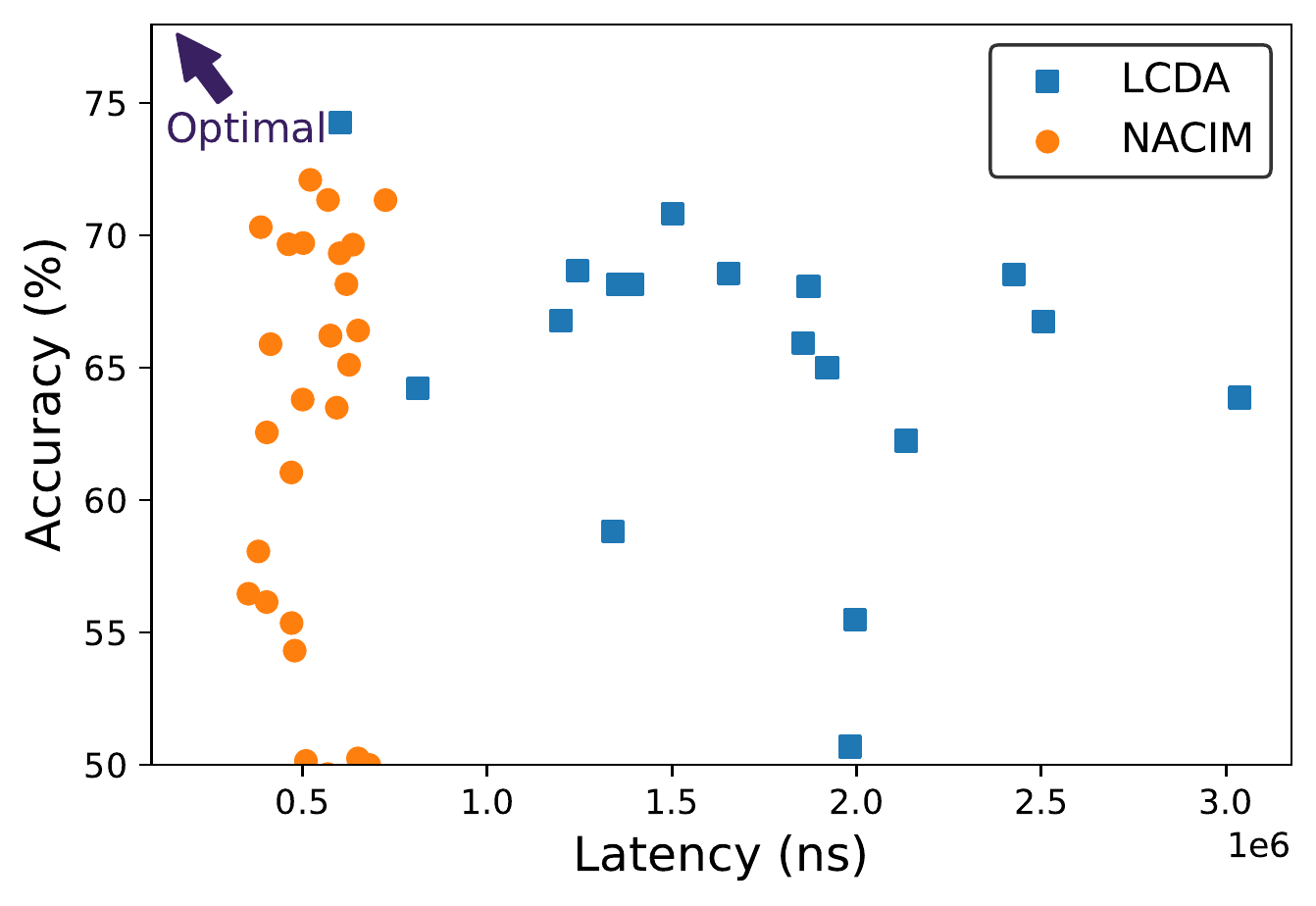}
    \vspace{-0.2cm}
    \caption{Accuracy-latency trade-offs of different design candidates provided by LCDA (blue square) and NACIM (orange dot). The Y-axis represents accuracy and the X-axis represents latency in ns.}
    \label{fig:sc_ltc}
    \vspace{-0.3cm}
\end{figure}

Fig.~\ref{fig:sc_ltc} presents the design candidates provided by LCDA and NACIM. Each data point denotes a design candidate; its Y-axis value indicates accuracy, and its X-axis value specifies latency (in ns). Designs nearer to the upper-left corner are more desirable.

In this scenario, however, LCDA falls short in providing designs that surpass those provided by NACIM, except for one outlier in the upper-left corner. LCDA struggles to deliver designs with sufficiently low latency.

Upon delving into each design candidate offered by LCDA, it became evident that LCDA's shortcomings stem from incorrect knowledge learned during GPT-4 pretraining. Specifically, GPT-4, having been primarily trained on deep learning papers, is deficient in understanding the behavior of DNN models deployed on CiM accelerators. It relies on its general knowledge, which can lead to errors when it conflicts with the unique requirements of CiM accelerators. 

Two fundamental misunderstandings by GPT-4 regarding kernel sizes stand out. First, while it's generally true that larger kernel sizes enhance accuracy, this doesn't always hold for CiM accelerators, given larger kernel sizes also increase the impact of device variations. Despite this, GPT-4 persistently enlarges kernel sizes to improve accuracy, leading to failures. Second, smaller kernel sizes typically imply lower latency in generic cases, but this is not universally valid. With kernel sizes of 3x3 and 7x7, the crossbar array is highly utilized, improving efficiency. However, a kernel size of 5x5 can result in a very low utilization rate and lower efficiency, potentially increasing latency. Unaware of this, GPT-4 consistently makes such mistakes.

These findings indicate that, in certain scenarios, the pre-trained GPT-4 model lacks the requisite knowledge for SW-HW co-design of CiM accelerators. A specific fine-tuning tailored to this task is necessary. Unfortunately, in this study, we don't have the privilege to fine-tune the GPT-4 model, hence we are unable to present results for a fine-tuned optimizer.

\subsection{Ablation Study: LLM Optimization without Knowledge}
We propose to involve LLMs in SW-HW co-design to solve the ``cold start'' problem. Here we show the result of using LLMs to optimize the designs without using carefully generated prompts that indicates that they are performing SW-HW co-design as an ablation study. This method is named LCDA-naive Here, we use trading off  energy consumption and accuracy as an example.
% The prompts are shown in Alg.~\ref{alg:random}.

% \begin{algorithm}
%     \caption{Random-Prompts~($l_{des}$, $l_{perf}$, \textit{Choices})}
%     \begin{algorithmic}\label{alg:random}
%         \STATE // Input: a list of explored design $l_{des}$, the corresponding normalized performance of each design $l_{perf}$, design space \textit{Choices}.
%         \STATE // Output: a prompt to the GPT model;
%         \STATE $prompt_s = $``\textit{You are a smart person.}'';
%         \STATE $prompt_u = $ ```
        
%         \textit{Your task is to assist me in selecting the best design in a search space. The design choices are:}
        
%         \textit{\{Choices\}}
        
%         \textit{Your objective is to define the optimal choice for each layer and maximize the reward. I will provide you with a reward.}
        
%         \textit{Your response should be the rollout list consisting of 6 number pairs(e.g. [[32,3],[32,3],[64,3],[64,3],[128,3],[128,3]]).}

%         \textit{Here are some experimental results that you can use as a reference:}
        
%         \{$l_{des}$, $l_{perf}$\}
        
%         \textit{Please suggest a rollout list that can improve the design's reward.}

%         \textit{Please do not include anything else other than the rollout list in your response.}
        
%         ''';
%         \STATE return $prompt_s$ + $prompt_u$;
%     \end{algorithmic}
% \end{algorithm}

\begin{figure}[ht]
    \centering
    \vspace{-0.2cm}
    \includegraphics[trim=0 0 0 0, clip, width=0.7\linewidth]{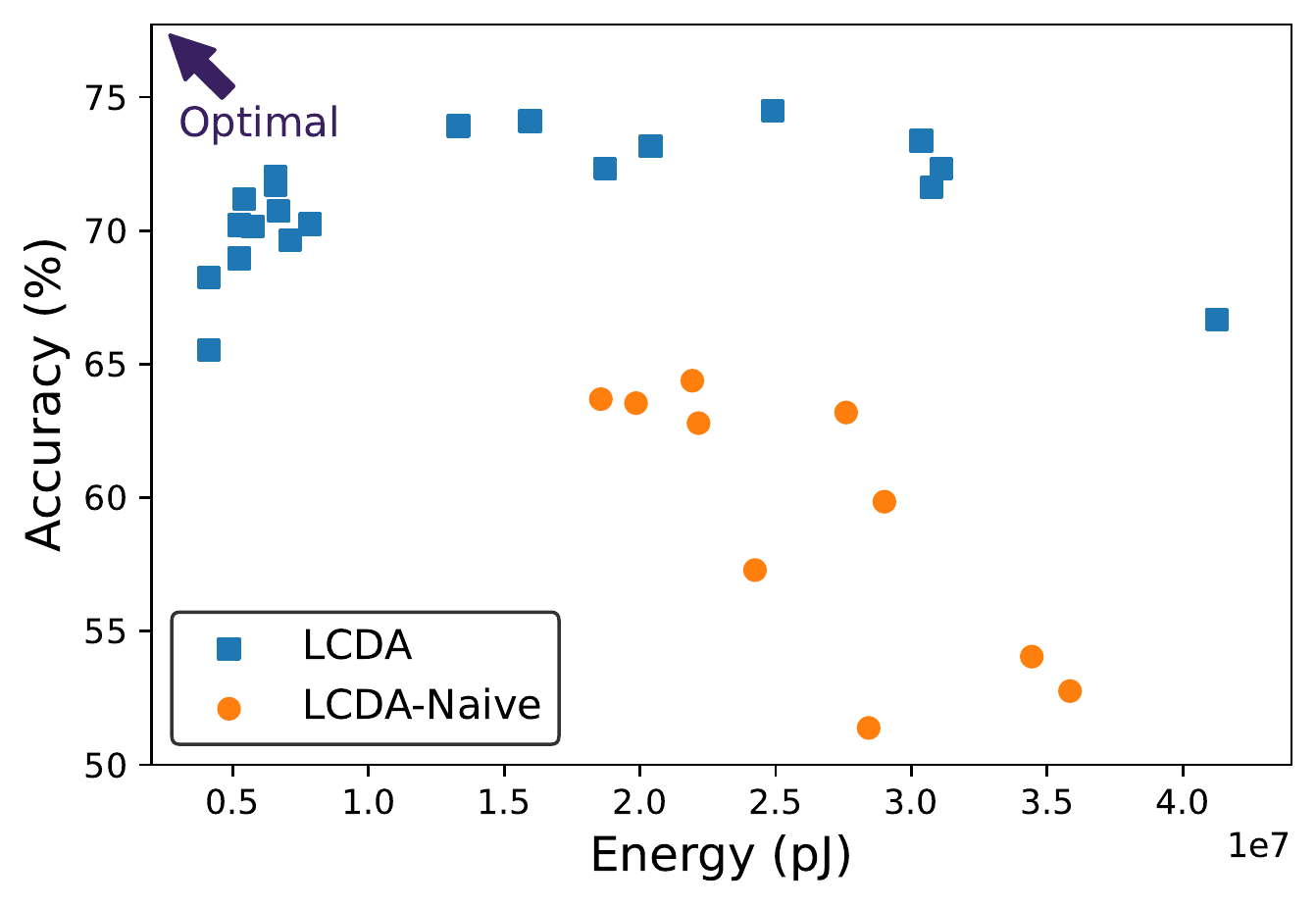}
    \vspace{-0.2cm}
    \caption{Accuracy-Energy trade-offs of different design candidates provided by LCDA (blue square) and LCDA-naive (orange dot). The Y-axis represents accuracy and the X-axis represents energy consumption in pJ.}
    \label{fig:sc_naive}
    \vspace{-0.3cm}
\end{figure}

Results in~\ref{fig:sc_naive} clearly show that, without knowing it's performing co-design tasks for DNN accelerators, LCDA-naive fails to provide efficient designs with optimal accuracy and energy consumption. This further demonstrates the importance of prior knowledge. 
\section{Conclusions}\label{sect:conclusion}

In this work, we propose the use of Language Models (LLMs) as optimizers for SW-HW co-design, as an alternative to RL or genetic algorithms. We introduce the LCDA framework to facilitate this approach. As an example task, we apply LCDA to co-designing DNN topology and CiM accelerators, demonstrating that LCDA achieves up to a 25x speedup while maintaining comparable performance to state-of-the-art co-design frameworks. 

Furthermore, we highlight some limitations of LCDA in certain corner cases, which pave the way for future research opportunities, including:

\begin{itemize}
    \item Explainable NAS: The changes in design parameters between consecutive episodes are human-readable, allowing users to request explanations by sending prompts to LLMs. This transparency breaks the black box nature of RL-based NAS and opens doors for leveraging the explainability of LLMs.
    \item Open-source LLMs: Our experiments reveal that optimal performance in co-design tasks often requires fine-tuning LLMs, which is not possible with commercial LLMs that function as black boxes. This presents opportunities for developing open-source LLMs that can be customized and tailored for co-design purposes.
    \item Design tools: While we hand-picked several hyperparameters and utilized NeuroSIM as our design tool, future exploration using AutoGPT~\cite{Auto-GPT} can unlock additional design tool possibilities and advancements in the field.
\end{itemize}

Overall, our work demonstrates the potential of leveraging LLMs and the LCDA framework for SW-HW co-design, while also highlighting directions for future investigations and advancements.

\bibliographystyle{ieeetr}
\bibliography{M7_References}

\end{document}